\documentclass[lettersize,journal]{IEEEtran}
\usepackage{amsmath,amsfonts}
\usepackage{array}
\usepackage[caption=false,font=normalsize,labelfont=sf,textfont=sf]{subfig}
\usepackage{textcomp}
\usepackage{stfloats}
\usepackage{url}
\usepackage{verbatim}
\usepackage{graphicx}
\def\BibTeX{{\rm B\kern-.05em{\sc i\kern-.025em b}\kern-.08em
    T\kern-.1667em\lower.7ex\hbox{E}\kern-.125emX}}
\usepackage{balance}

\usepackage{tcolorbox}

\newtcolorbox{reasonbox}{
  colback=gray!5,  
  colframe=gray!50!black,  
  fonttitle=\bfseries,
  title=Reason,
  arc=0mm,  
  boxrule=0.5pt  
}

\usepackage{courier} 
\newtcolorbox{promptbox}{
  colback=gray!5,  
  colframe=gray!50!black,  
  fonttitle=\fontfamily{pcr}\selectfont, 
  title=Prompt,
  arc=0mm,  
  boxrule=0.5pt  
}
\usepackage{listings}
\definecolor{codegreen}{rgb}{0,0.6,0}
\definecolor{codegray}{rgb}{0.5,0.5,0.5}
\definecolor{codepurple}{rgb}{0.58,0,0.82}
\definecolor{backcolour}{rgb}{0.95,0.95,0.92}

\lstdefinestyle{mystyle}{
  commentstyle= \color{red!50!green!50!blue!50},  
  keywordstyle= \color{blue!70},  
  stringstyle=\rmfamily,
  basicstyle=\rmfamily\footnotesize,
  columns=[c]flexible,      
  basewidth={0.4em,0.4em},
  breakatwhitespace=false,
  breaklines=true,  
  breakindent=0pt,      
  captionpos=b,
  keepspaces=true,
  showspaces=false,
  showstringspaces=false,  
  showtabs=false,
  tabsize=2,
  frame=single  
}
\usepackage{color}

\usepackage{algorithm}
\usepackage{algpseudocode}
\usepackage{multirow}

\begin{document}
\title{FashionM3: Multimodal, Multitask, and Multiround Fashion Assistant based on Unified Vision-Language Model}
\author{
\IEEEauthorblockN{Kaicheng Pang$^{1,2}$, Xingxing Zou$^{2}$, Waikeung Wong$^{1,2}$ \\}
\IEEEauthorblockA{
$^{1}$Laboratory for Artificial Intelligence in Design, Hong Kong SAR, China\\
$^{2}$School of Fashion and Textiles, Hong Kong Polytechnic University, Hong Kong SAR, China}
\thanks{$^*$Waikeung Wong is the corresponding author.}
}

\markboth{IEEE Transactions on Neural Networks and Learning Systems}%
{Manuscript}

\maketitle

\begin{abstract}
Fashion styling and personalized recommendations are pivotal in modern retail, contributing substantial economic value in the fashion industry. With the advent of vision-language models (VLM), new opportunities have emerged to enhance retailing through natural language and visual interactions. This work proposes FashionM3, a multimodal, multitask, and multiround fashion assistant, built upon a VLM fine-tuned for fashion-specific tasks. It helps users discover satisfying outfits by offering multiple capabilities including personalized recommendation, alternative suggestion, product image generation, and virtual try-on simulation. Fine-tuned on the novel FashionRec dataset, comprising 331,124 multimodal dialogue samples across basic, personalized, and alternative recommendation tasks, FashionM3 delivers contextually personalized suggestions with iterative refinement through multiround interactions. Quantitative and qualitative evaluations, alongside user studies, demonstrate FashionM3's superior performance in recommendation effectiveness and practical value as a fashion assistant.
\end{abstract}


\section{Introduction}
Clothing is a powerful medium for expressing personal identity and style~\cite{saha2024comprehensive}. The fashion industry increasingly relies on recommendation systems to curate personalized outfits, as traditional and e-commerce approaches often struggle to balance expertise, customer preferences, and real-time adaptability~\cite{burke2002hybrid,grewal2018evolution,rana2024knowledge,qu2024assembled,kwon2024future}. To address these challenges, numerous approaches~\cite{kaicheng2021modeling,ranjan2024fashion,su2024personalized,tosun2024impact,shankar2024intelligent} have been proposed for fashion recommendation. Ranking-based methods~\cite{lu2019learning,tan2019learning,lin2020fashion,vasileva2018learning} focus on learning the representations of fashion items and modeling the compatibility among items in an outfit. During the recommendation phase, they generate a set of outfit candidates and rank them based on a score that represents the compatibility of the outfit or user preferences. However, these methods face significant practical limitations. The main challenge lies in the combinatorial explosion of possible outfit combinations, which makes it computationally infeasible to evaluate all potential candidates when dealing with extensive inventories. These methods also fail to deliver context-aware recommendations based on real-time user queries.

With the advancement of powerful generative models~\cite{brown2020language, dhariwal2021diffusion}, fashion recommendation approaches have undergone a paradigm shift from ranking-based methods to generation-based approaches~\cite{moosaei2022outfitgan, xu2024diffusion, wang2023generative, shi2025integrating}, enabling direct suggestions tailored to users. On one hand, Shi \textit{et al.}~\cite{shi2025integrating} developed a language-based model, which employs auto-prompt generation and retrieval augmentation to deliver fashion recommendations in natural language. On the other hand, Moosaei \textit{et al.}~\cite{moosaei2022outfitgan} proposed a image generation model combining a Generative Adversarial Network to create realistic fashion items with a Compatibility Network to ensure stylistic coherence among outfit components. Xu \textit{et al.}~\cite{xu2024diffusion} introduced DiFashion, leveraging diffusion models to generate fashion images based on historical and mutual compatibility conditions.

However, these methods lack multimodal integration, limiting their effectiveness in providing comprehensive fashion recommendations. The image-based approaches of ~\cite{moosaei2022outfitgan} and ~\cite{xu2024diffusion} cannot produce descriptive explanations or incorporate textual user inputs, such as detailed preferences or occasion-specific requirements. Conversely, the language-based model of ~\cite{shi2025integrating} lacks the capability to generate images of recommended items. Additionally, none of these methods effectively combine real-time user feedback with contextual factors like personal style or alternative preference, hindering their ability to deliver dynamic, contextually personalized recommendations.

To address the limitations above, we propose a Multimodal, Multitask, and Multiround Fashion Assistant (FashionM3) built upon a unified VLM to enable interactive fashion recommendation. FashionM3 can simultaneously process a user's \textbf{multimodal} query, enabling users to initiate queries through natural language conversations and image uploads, thus capturing diverse contextual factors. The \textbf{multitask} capability allows FashionM3 to handle various fashion-related tasks, including outfit completion, personalized recommendation, alternative suggestion, fashion image generation, and virtual try-on simulation, enabling it to address multiple user needs within a single framework and bypass the combinatorial explosion problem by directly generating contextually relevant suggestions. Users can provide feedback through \textbf{multiround} interactions, enabling FashionM3 to iteratively refine suggestions via continuous dialogue.

At the core of FashionM3, we develop FashionVLM, a VLM fine-tuned for fashion-specific tasks, integrating fashion aesthetics and image generation capabilities. To enable training, we introduce a novel dataset named FashionRec\footnote{Available at \url{https://huggingface.co/datasets/Anony100/FashionRec}}, including 331,124 image-dialogue pairs spanning on three key types of recommendation tasks: Basic Recommendation, Personalized Recommendation, and Alternative Recommendation. These training samples enable FashionVLM to learn fashion aesthetic logic, rationales for recommendations, and effective incorporation of user preferences. We propose a pipeline to transform manually curated outfits and user interaction histories included in existing fashion datasets into fashion recommendation dialogues. We develop the FashionM3 system based on the Model Context Protocol (MCP) to integrate multiple components, namely the FashionVLM for fashion recommendation, the User Database for retrieving user interaction records, the Product Database for retrieving product information, and virtual try-on for outfit simulation.

All in all, our contributions can be summarized as follows:

\begin{itemize}
    \item We introduce FashionM3, a pioneering fashion assistant that leverages a unified VLM to deliver fashion recommendation and image generation. It supports multimodal inputs, performs multiple fashion-related tasks, and enables multi-round interactions for dynamic refinement of recommendations.
    \item We introduce the FashionRec dataset, where a novel pipeline is developed that can transform manually curated outfits and user interaction histories into high-quality fashion recommendation dialogues.
    \item We present both quantitative and qualitative evaluations to validate the effectiveness of our fine-tuning approach, demonstrating that FashionVLM’s recommendation performance surpasses baselines. Additionally, we perform user studies to showcase FashionM3’s practical value, highlighting its effectiveness in delivering satisfying fashion advice to users.
\end{itemize}

\section{Related Work}
\subsection{Fashion Recommendation}
With the continuous expansion of e-commerce, fashion recommendation has attracted significant research attention~\cite{chakraborty2021fashion, 10.1145/3624733,ranjan2024fashion,su2024personalized,tosun2024impact,shankar2024intelligent,kalinin2024generative,herzallah2024fashioning}. The development of outfit recommendation methods has progressed through three stages based on the information used in the scoring function. Specifically, the first stage focuses on predicting fashion compatibility by learning intrinsic relationships among fashion items~\cite{mcauley2015image,10049142, kaicheng2021modeling, tangseng2020toward}. Subsequently, researchers~\cite{hu2015collaborative, lu2021personalized, chen2019pog, li2020hierarchical} utilized collaborative filtering to model user preferences based on user interaction data for personalized recommendation problem. The third stage incorporates contextual information to better understand users' needs, where researchers consider either user characteristics (\textit{e.g.}, physical attributes~\cite{pang2024learning,hidayati2018dress}, hair-style~\cite{yang2012hairstyle, liu2014wow}) or environmental factors (\textit{e.g.}, occasion~\cite{ye2023show}, weather~\cite{adewumi2020unified}) to deliver more targeted recommendations.
In practical applications, existing outfit recommendation systems~\cite{lin2019explainable,lu2021personalized,bettaney2021fashion,kaicheng2021modeling} typically operate as ranking-based methods. They first construct a candidate set $\Omega$ from the inventory and then select the optimal outfit through scoring and ranking.

\[O^* = \mathop{\arg\max}_{O \in \Omega} S(O)\]
where $O^*$ represents the optimal outfit from the candidate outfit set $\Omega$, and $S(\cdot)$ denotes the scoring function. The solution space $\Omega$ can be defined as:

\[\Omega = \{O | O \subseteq I, g(O) \leq 0\}\]
where $I = \{i_1, i_2, ..., i_n\}$ denotes the fashion item inventory, and $g(O)$ represents the constraint function. Unlike the common recommendation problem for a single product, the solution space size for outfit recommendation $\Omega$ grows combinatorially with the inventory size $n$. This combinatorial complexity consequently renders ranking-based methods computationally impractical for evaluating all possible outfits, particularly on large-scale e-commerce platforms (\textit{e.g.}, over a million fashion items on Farfetch.com). While sampling strategies\cite{chen2019pog, dong2020fashion} have been proposed to reduce the search space; however, they often lead to suboptimal recommendations due to the lack of semantic understanding of fashion compatibility\cite{song2020knowledge}.

These limitations have prompted the exploration of generation-based approaches for fashion recommendation. OutfitTransformer~\cite{sarkar2022outfittransformer} pioneered this direction by reformulating outfit compatibility prediction and complementary item retrieval as sequence-generation tasks through task-specific tokens and self-attention mechanisms. DiFashion~\cite{xu2024diffusion} advanced this approach by introducing generative outfit recommendations that leverage diffusion models to generate visually compatible fashion items tailored to user preferences. Ma \textit{et al.}~\cite{ma2024leveraging} proposed a contrastive learning-enhanced hierarchical encoder to generate bundle recommendations by unifying multimodal features and item-level user feedback. However, these approaches struggle to effectively integrate multimodal inputs to handle diverse user requests.

Meanwhile, recent advances in conversational fashion recommendation systems have focused on improving user interaction mechanisms. Ye \textit{et al.}~\cite{yeAIYoEmbeddingPsychosocial2024} incorporated psychosocial aspects into chatbot design, enabling users to express preferences through a list-based interaction mechanism. Wu \textit{et al.}~\cite{wu2022multimodal} investigated the effectiveness of positive and negative natural language feedback in multimodal conversational recommendation. At the same time, their subsequent work~\cite{wu2022multi} proposed a dialog state-tracking framework to capture user preferences through multi-turn interactions. However, these conversational systems are primarily designed for single-task scenarios or rely heavily on pre-defined interaction patterns, limiting their ability to handle diverse fashion-related queries and provide comprehensive styling advice. 

To overcome these limitations, we propose FashionM3, a multimodal fashion assistant built on a VLM fine-tuned for fashion-specific tasks. Unlike ranking-based methods, FashionM3 bypasses combinatorial complexity by directly generating contextually relevant outfits, trained on our novel FashionRec dataset to ensure robust semantic understanding of fashion compatibility. It effectively integrates multimodal inputs, such as text queries and image uploads, to address diverse user requests, including personalized recommendations and style preferences. Additionally, FashionM3’s multitask and multiround capabilities enable it to handle various fashion tasks while supporting dynamic user interactions, overcoming the single-task constraints of prior conversational systems.

\vspace{-3mm}
\subsection{Unified Vision-Language Model}
Unified VLM have become a hot research topic that can handle understanding and generation tasks within a single architecture. Xie \textit{et al.}~\cite{xie2024show} introduced Show-O that combines autoregressive and discrete diffusion modeling, enabling flexible handling of various vision-language tasks from visual question-answering to text-to-image generation. JanusFlow~\cite{ma2024janusflow} presented a minimalist architecture that integrates autoregressive language models with rectified flow. This demonstrates that unified training can be achieved without complex architectural modifications by decoupling understanding and generation encoders while aligning their representations. Fei \textit{et al.}~\cite{fei2024vitron} proposed VITRON, a universal pixel-level vision LLM that supports comprehensive understanding, generating, segmenting, and editing of visual content through a hybrid approach that integrates discrete textual instructions with continuous signal embeddings. In the fashion domain, UniFashion~\cite{zhao2024unifashion} unifies retrieval and generation tasks by integrating diffusion and large language models. Nevertheless, these models focus on general-purpose tasks, lacking specialized capabilities for fashion understanding. To address this gap, we fine-tune the Show-O model on our novel FashionRec dataset to enhance its multimodal understanding and image generation ability tailored to fashion recommendation tasks.

\section{FashionRec Dataset}
\label{sec:dataset}
\begin{figure*}[t]
  \includegraphics[width=\textwidth]{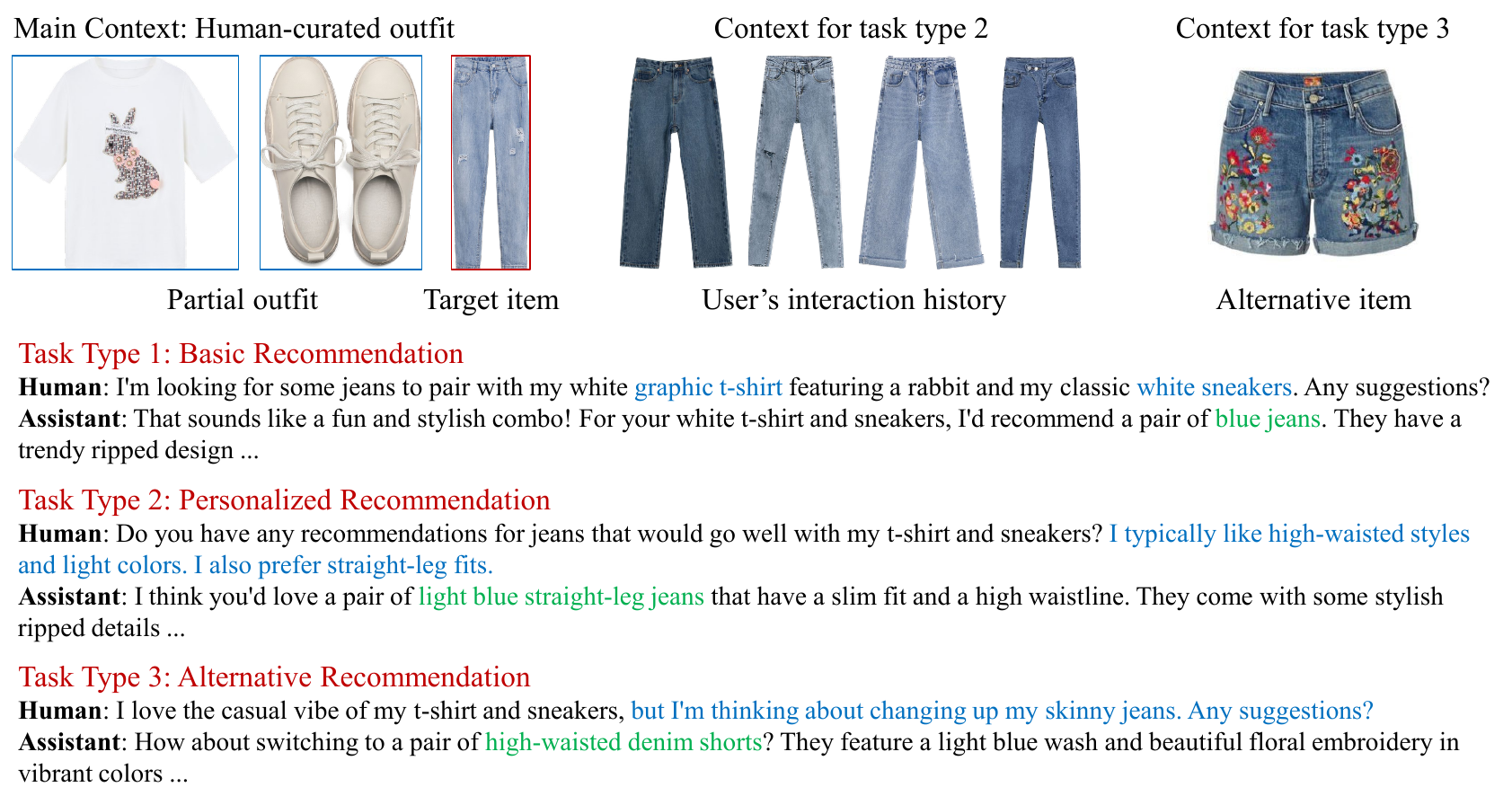}
  \vspace{-7mm}
  \caption{Three examples from the FashionRec dataset illustrate its three task types. Blue and green text highlight key elements of the human and assistant dialogue, respectively, generated by a large language model based on provided outfit and recommendation scenarios. Basic recommendation suggests a target item (\textit{e.g.}, blue jeans) to complete a partial outfit (\textit{e.g.}, t-shirt and sneakers), both derived from the same human-curated outfit. Personalized recommendation incorporates the user's interaction history (\textit{e.g.}, user prefers light colors) to delieve personalized recommendations (light blue straight-leg jeans). Alternative recommendation suggests an alternative item (\textit{e.g.}, denim shorts) to replace a fashion piece of the same category (\textit{e.g.}, pants) in an outfit, using two human-curated outfits sharing at least two common items.}
  \vspace{-4mm}
  \label{fig:dataset}
\end{figure*}

While numerous fashion datasets~\cite{zou2019fashionai,lu2019learning,lai2019theme} are available, they only feature human-curated outfit combinations and lack the image-dialogue data pairs necessary to train VLM for fashion recommendation. To bridge this gap, we propose the FashionRec (Fashion Recommendation) Dataset, a multimodal dataset specifically crafted to enable the knowledge distillation of a pre-trained VLM. This dataset integrates aesthetic coherence from human-curated outfits with personalized user preferences derived from interaction histories. FashionRec comprises 331,124 fashion recommendation samples spanning three key types of recommendation tasks, as illustrated in Fig.~\ref{fig:dataset}. The basic recommendation task focuses on suggesting items to complete a partial outfit. The second task is personalized recommendation, which tailors suggestions based on the user's interaction history. As for the alternative recommendation task, it proposes variations to an outfit by swapping items.

\begin{table}[t]
  \centering
  \caption{Statistics of the FashionRec dataset by source. `-' denotes that the corresponding information is not available.}
  \vspace{-2mm}
  \label{table:dataset-statistics}
  \resizebox{0.5\textwidth}{!}{
  \begin{tabular}{l|ccccc}
      \hline
      Source & \#Item & \#Outfit & \#Items/Outfit & \#User & \#Outfits/User\\
      \hline
      iFashion  & 37,018 & 29,739 & 3.86 & 2,299 & 88.77\\
      Polyvore-519 & 102,148 & 59,563 & 3.58 & 519 & 114.79\\
      Fashion32 & 26,925 & 13,981 & 2.91 & - & -\\
      \hline
      Total & 166,091 & 103,283 & 3.57 & 2,818 & 93.56 \\
      \hline
  \end{tabular}
  }
  \vspace{-3mm}
\end{table}

\begin{table}[t]
\centering
  \caption{Statistics of the FashionRec dataset splits for each task.}
  \vspace{-2mm}
  \label{table:dataset-split}
  \begin{tabular}{l|ccc}
      \hline
      Task & Train & Valid & Test \\
      \hline
      Basic Recommendation & 86,776 & 6,763 & 9,741 \\
      Personalized Recommendation & 208,599 & 3,904 & 5,320 \\
      Alternative Recommendation & 8,559 & 743 & 719 \\
      \hline
  \end{tabular}
  \vspace{-5mm}
\end{table}

We construct the FashionRec dataset by sourcing data from three wildly used fashion datasets: iFashion~\cite{chen2019pog}, Polyvore-519~\cite{lu2019learning}, and Fashion32~\cite{lai2019theme}, as detailed in Table~\ref{table:dataset-statistics}. 
The FashionRec dataset comprises 103,283 outfits, including 166,091 items, contributed by 2,818 users. The training split includes 86,776 samples for Basic Recommendation, 208,599 for Personalized Recommendation, and 8,559 for Alternative Recommendation, with their validation and test splits shown in Table~\ref{table:dataset-split}. These samples enable FashionM3 to learn fashion aesthetic principles, recommendation rationales, and effective user preference integration. Prompts used to generate these conversations are presented in Appendix~\ref{appendix:recommendation-based-queries} and more examples can be found in Appendix~\ref{appendix:dataset-examples}. We now describe the methodology for constructing the image-dialogue pairs in the FashionRec dataset.

\noindent\textbf{Basic Recommendation.}
The foundation of the FashionRec Dataset rests on human-assembled fashion outfits. Consider a user who has curated an outfit $O$, comprising $n$ items denoted as $O = \{i_1, i_2, ..., i_n\}$. Each item $i$ is associated with an image $v$ and a textual description $t$. We assume $O$ is aesthetically coherent due to its human origin. To generate basic recommendation dialogue, we decompose $O$ into two disjoint subsets: \textbf{Partial Outfit}: $P \subset O$ and \textbf{Target Item Set}: $T = O \setminus P$, the remaining items to be predicted given $P$, as illustrated in the top-right corner of Fig.~\ref{fig:dataset}. The way to construct basic recommendation data is to ask Large Language Model (LLM) to generate queries $X_q$ and responses $X_a$ based on $P$ as the input condition and $T$ as the target items, using their textual descriptions. It is worth noting that $T$ may include several items, thus, multiple conversations are generated for each item in $T$ to ensure diversity. During training, the model needs to predict all responses in $X_a$.

\begin{algorithm}[t]
\caption{User History Filtering Algorithm}
\label{alg:user-outfit}
\begin{algorithmic}[1]
\State \textbf{Input:} Outfit $O = \{i_1, i_2, ..., i_n\}$, User $u$, Minimum item interactions $m_i$, Minimum user history interactions $m_u$
\State \textbf{Output:} Partial outfit $P$, Target item $t$, Filtered user history $U_c'$

\State $R \gets \emptyset$ \Comment{Initialize empty set for possible results}

\For{each $i \in O$} \Comment{Iterate through all items in $O$}
  \State $P \gets O \setminus \{i\}$ \Comment{Partial outfit}
  \State $c \gets$ category of $i$
  \State $U_c \gets$ get items in category $c$ user interacted with
  \State $H_c \gets$ get items in category $c$ co-occurred with $P$ \Comment{Items compatible with $P$}

  \If {$\| U_c \| \ge m_u \wedge \| H_c \| \ge m_i$} 
    \State Add $\{P, i, H_c, U_c\}$ to $R$
  \EndIf
\EndFor

\If {$R = \emptyset$}
  \State \textbf{return} $\mathrm{None}$
\EndIf

\State $r^* \gets \arg\max_{r \in R} (\alpha \times \| r.H_c \| + \| r.U_c \|)$
\State $P \gets r^*.P$, $t \gets r^*.i, H_c \gets r^*.H_c, U_c \gets r^*.U_c$ \Comment{Get optimal partial outfit and target item pair}
\For{each $j \in H_c$}
  \State $e_v \gets$ CLIP image embedding of $v_j$
  \State $e_t \gets$ CLIP text embedding of $t_j$
  \State $f_j \gets (e_v + e_t) / 2$   \Comment{Get item feature vector}

  \State $\text{count}(j) \gets$ interaction count of item $j$ with $P$
  \State $w_j \gets \text{count}(j) / \sum_{k \in H_c} \text{count}(k)$
\EndFor
\State $f_{H_c} \gets \sum_{j \in H_c} (w_j \cdot f_j)$ \Comment{Weighted average feature vector of compatible items}
\For{each $j \in U_c$}
  \State $\text{sim}(j) \gets \frac{f_j \cdot f_{H_c}}{\|f_j\| \|f_{H_c}\|}$ \Comment{Cosine similarity}
  \State $\text{count}(j) \gets$ interaction count of item $j$ with $u$
  \State $\text{score}(j) \gets (\beta \cdot \frac{\text{count}(j)}{\max_{k \in U_c} \text{count}(k)} + 1) \cdot \text{sim}(j) $
\EndFor
\State $U_c' \gets$ top-$k$ items in $U_c$ ranked by $\text{score}$ in descending order

\State \textbf{return} $P$, $t$, $U_c'$
\end{algorithmic}
\end{algorithm}

\noindent\textbf{Personalized Recommendation.}
We leverage the user's interaction histories as context to construct personalized recommendation dialogues.
Specifically, for a target item $i \in T$ belonging to category $c$, we define $U_c$ as the set of historical items in category $c$ with which user $u$ has interacted. Given that $\left | U_c \right | $ can be considerable (average of 17.15 items per category per user for Polyvore-519 dataset) and exhibits significant stylistic diversity within the same category, directly incorporating all interacted items into the training and inference stages is inappropriate. 
To address this, we filter users' previously interacted items during the data preparing process, as detailed in the Algorithm~\ref{alg:user-outfit}. The goal is to filter $U_c$ into a refined subset $U_c'$ based on fashion compatibility with the partial outfit and interaction frequencies. The process is outlined as follows.

We first decompose the outfit $O$ by selecting each item $i \in O$ as a target item $t$, with the remaining items forming a partial outfit $P$, thus creating $\left | O \right | $ candidate pairs $\{P, t \}$. For each pair, we identify $U_c$ as the user's historical items in category $c$ and $H_c$ as items in category $c$ that co-occur with $P$. We filter these pairs to form a candidate set $R$, retaining only those where $U_c$ and $H_c$ meet thresholds $m_u =10$ and $m_i = 3$, ensuring $\| U_c \| \ge 10 \wedge \| H_c \| \ge 3$ to provide sufficient data for recommendation. These values are set based on empirical experience. 43\% of the data is excluded with these thresholds, striking a balance between ensuring historical interactions and compatibility evidence while preventing overly restrictive constraints.

From the candidate set $R$, we select the optimal pair $r^*$ by maximizing $\alpha \times \| r.H_c \| + \| r.U_c \| $, where $\alpha = 3$. This weighting reflects the observed data distribution, with a mean size ratio of $U_c$ to $H_c$ of approximately 1.65, but we set $\alpha = 3$ to emphasize outfit coherence. Next, we compute feature vectors for items in $H_c$ using CLIP model~\cite{radford2021learning}, averaging image and text embeddings as $f_j = (e_v + e_t) / 2$. The weighted average feature vector $f_{H_c} \gets \sum_{j \in H_c} (w_j \cdot f_j)$ is calculated with weights based on interaction counts with $P$.
Finally, for each item $j \in U_c$, we calculate a similarity score using cosine similarity, $\text{sim}(j) \gets \frac{f_j \cdot f_{H_c}}{\|f_j\| \|f_{H_c}\|}$, and combine it with interaction frequency to compute the final score:
\begin{equation}
  \text{score}(j) = \left( \beta \cdot \frac{\text{count}(j)}{\max_{k \in U_c} \text{count}(k)} + 1 \right) \cdot \text{sim}(j),
\end{equation}
where $\beta = 2.0$. Here, count($j$) is the interaction count of item $j$ with user $u$, normalized by the maximum count in $U_c$ to scale its influence. The term $\beta \cdot \frac{\text{count}(j)}{\max_{k \in U_c} \text{count}(k)}$ amplifies the contribution of frequently interacted items. At the same time, the constant 1 ensures that similarity remains a baseline factor even for items with low interaction counts. The top-$k$ items in $U_c$, ranked by score($j$) in descending order, form the filtered set $U_c'$.
We asked LLM to utilize the partial outfit $P$, target item set $T$, and filtered user history $U_c'$ as input to generate user queries $X_q$ and corresponding responses $X_a$.

\noindent\textbf{Alternative Recommendation.}
The alternative recommendation task supports scenarios where users seek to modify an existing outfit while preserving its aesthetic coherence. For a given outfit $O_a = \{ i_1, i_2, \dots, i_m \}$, we identify another outfit $O_b = \{ j_1, j_2, \dots, j_n \}$ such that the two outfits share at least two identical items. The overlapping items, denoted as $S = O_a \cap O_b$, are anchors to maintain outfit consistency. The replaceable items are drawn from the non-overlapping sets $O_a \setminus S$ and $O_b \setminus S$. Specifically, for an item $i_a \in O_a \setminus S$ that the user wishes to replace, we select an alternative $i_b \in O_b \setminus S$ where $i_a$ and $i_b$ belong to the same category, ensuring the replacement fits the same role within the outfit. To generate dialogue data, $O_a$ is treated as the user's current outfit, with $S$ as the partial outfit and $i_a$ as the item to be replaced. We then prompt LLM to generate conversations suggesting $i_b$ to replace $i_a$, given the input outfit $O_a$.

\section{FashionM3}
\subsection{Training of Fashion Vision-Language Model}
\begin{figure}[t]
  \includegraphics[width=0.5\textwidth]{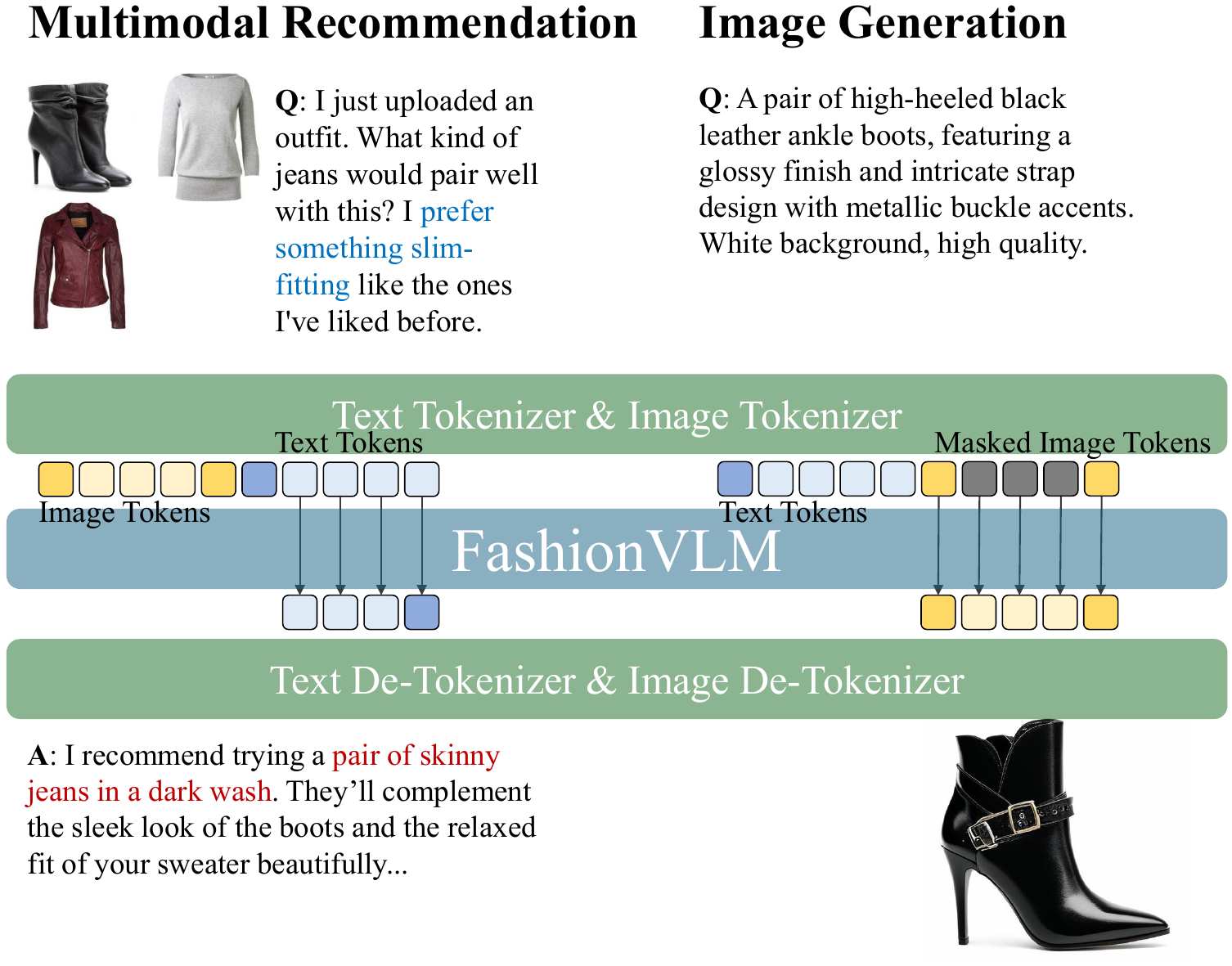}
  \vspace{-7mm}
  \caption{Training pipeline of FashionVLM, showcasing the multimodal recommendation and text-to-image generation tasks. In the multimodal recommendation task, a user provides an outfit and a preference for jeans, receiving a recommendation for dark-wash skinny jeans. In the text-to-image generation task, the model generates an image of high-heeled black leather ankle boots based on a provided textual description. Inputs are processed by a text tokenizer and an image tokenizer, which convert queries and images into text and image tokens for FashionVLM. Outputs are then transformed back into textual and visual formats using a text de-tokenizer and an image de-tokenizer.}
  \vspace{-5mm}
  \label{fig:fashionVLM}
\end{figure}

To address the limitations of existing multimodal models, which often fall short in fashion-specific contexts, we train FashionVLM as the core component of FashionM3. The training pipeline of FashionVLM is depicted in Fig.~\ref{fig:fashionVLM}. We utilize Show-O~\cite{xie2024show} as its model architectural and fine-tune it on the FashionRec dataset to tailor it for fashion-specific queries. Below, we detail two training tasks and their objectives.

\noindent Multimodal recommendation task trains FashionVLM to generate a textual response $\hat{X}_a$ containing recommendation information, conditioned on the input image and the user query. The input consists of fashion item images $X_v$, user query $X_q$, and the target response $X_a$. Specifically, $X_v$ is encoded into a sequence of visual tokens $\{u_i\}_{i=1}^M$ using a pre-trained look-up-free quantizer MAGVIT-v2~\cite{yu2023language}. Both the user query $X_q$ and the target response $X_a$ are tokenized using the phi model's~\cite{gunasekar2023textbooks} tokenizer, producing textual token sequences $\{v_i^q\}_{i=1}^{N_q}$ and $\{v_i^a\}_{i=1}^{N_a}$, respectively. During training, the model takes the visual tokens $\{u_i\}_{i=1}^M$ and the query tokens $\{v_i^q\}_{i=1}^{N_q}$ as input and is tasked with predicting the response tokens $\{v_i^a\}_{i=1}^{N_a}$. The training objective is formulated as a next-token prediction loss, computed solely over the tokens of $X_a$:
\begin{equation}
  \mathcal{L}_{\text{MMR}} = -\sum_{i=1}^{N_a} \log P_\theta(v_{i+1}^a | u_1, \dots, u_M, v_1^q, \dots, v_{N_q}^q, v_1^a, \dots, v_i^a),
\end{equation}
where $P(v_{i+1}^a | v_1^q, \dots, v_{N_q}^q, u_1, \dots, u_M, v_1^a, \dots, v_i^a)$ denotes the conditional probability of the next token $v_{i+1}^a$ in the response sequence, given the query tokens $\{v_1^q, \dots, v_{N_q}^q\}$, the visual tokens $\{u_1, \dots, u_M\}$, and the preceding response tokens $\{v_1^a, \dots, v_i^a\}$. $\theta$ represents the parameters of FashionVLM. This objective ensures that FashionVLM generates responses $X_a$ that are contextually relevant to the user's query $X_q$ and informed by the visual content in $X_v$.

\noindent Text-to-image generation focuses on generating visual tokens for images based on textual descriptions. We adopt a discrete diffusion-based approach for modeling the image tokens $\{u_i\}_{i=1}^M$. During training, a subset of the image tokens is randomly replaced with a special \textit{[MASK]} token at a ratio determined by a time step, creating a masked sequence $\{u_i^*\}_{i=1}^M$, where $u_i^* = u_i$ if the $i$-th token is unmasked, and $u_i^* = \textit{[MASK]}$ if masked. The model is then tasked with reconstructing the original image tokens at the masked positions, conditioned on the entire visual sequence (including both unmasked and masked tokens) and the textual tokens $\{v_i\}_{i=1}^N$ encoded from the input description $X_q$. The training objective is to maximize the likelihood of predicting the masked tokens, formulated as a mask token prediction loss:

\begin{equation}
  \mathcal{L}_{\text{T2I}} = -\sum_{j \in \mathcal{M}} \log P_\theta(u_j | \{u_i^*\}_{i=1}^M, v_1, v_2, \dots, v_N),
\end{equation}

where $\mathcal{M}$ denotes the set of indices of the masked tokens, and $P_\theta(u_j | \{u_i^*\}_{i=1}^M, v_1, v_2, \dots, v_N)$ represents the probability of reconstructing the original visual token $u_j$ at a masked position $j \in \mathcal{M}$, conditioned on the entire visual and textual tokens. The loss is computed exclusively over the masked positions, ensuring the model learns to generate coherent visual tokens by leveraging the unmasked visual context and the textual input.

\begin{figure}
  \includegraphics[width=0.5\textwidth]{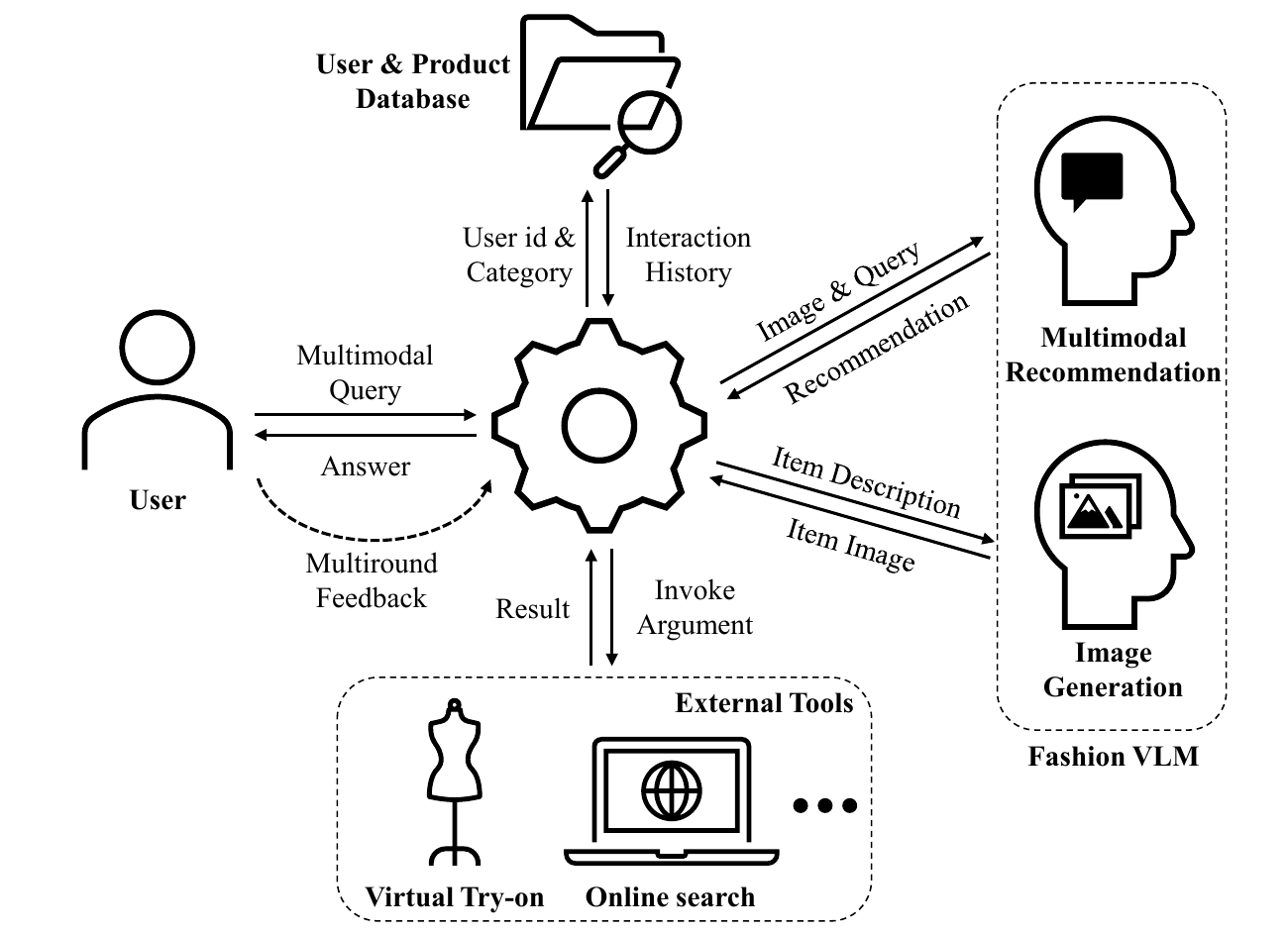}
  \vspace{-7mm}
  \caption{Overview of FashionM3's architecture, orchestrating the flow of information across key components: multimodal query understanding, user data retrieval, FashionVLM for recommendation and image generation, and integration of external tools.}
  \vspace{-6mm}
  \label{fig:flow}
\end{figure}
\vspace{-2mm}
\subsection{System Architecture of FashionM3}
\label{section:architecture}
Built upon FashionVLM, we develop FashionM3, a user-centric recommendation framework designed to deliver personalized and interactive recommendation experiences. As depicted in Fig.~\ref{fig:flow}, the architecture leverages the Model Context Protocol (MCP) client as the central processing unit to orchestrate the flow of information across distinct stages. The process starts with users submitting multimodal queries. These queries are refined through multiround feedback to capture their intent. The OpenAI GPT-4o model then processes these inputs to interpret the query and invoke appropriate tools. User-specific information, including profiles, preferences, and interaction histories, is retrieved from the User and Product Database, which is encapsulated as an MCP server. FashionVLM, integrated as an MCP server, generates personalized recommendation contents using refined queries and summarized user preferences. It can also generate product images via the Image Generation module, if requested. External tools like Virtual Try-on and Online Search are also integrated as MCP servers to enrich functionality. Finally, the response generation stage delivers multimodal recommendations that combine visual and textual outputs. Built with the Chainlit~\cite{chainlit} framework, FashionM3 offers an intuitive conversational AI interface. Its modular MCP-based design enables future integrations, such as e-commerce payment processing, to support the complete user journey from recommendations to purchase within a streamlined workflow. The system operates efficiently on a server with a single NVIDIA 4090D GPU.

\section{Experiments}
This section presents a comprehensive evaluation of FashionM3 through a quantitative assessment using the FashionRec dataset's test split, a qualitative recommendation performance comparison with baselines, an in-depth user study demonstrating its practical value in real-world fashion recommendation scenarios, and a detailed analysis of usage scenarios.

\vspace{-4mm}
\subsection{Experimental Settings}
\noindent\textbf{Datasets.} We utilize the test split of FashionRec datasets to compare recommendation performance with baselines. Three tasks are involved: basic recommendation, personalized recommendation, and alternative recommendation.

\vspace{2mm}

\noindent\textbf{Baselines.} We compare our proposed method with several off-the-shelf VLMs.
1) \textbf{Show-O}~\cite{xie2024show} is a multimodal VLM designed to handle visual understanding and image-generation tasks in a unified manner. We directly compare Show-O with FashionVLM to validate the fine-tuning effects.
2) \textbf{LLaMA-3.2-vision}~\cite{meta2023llama32} is pretrained and instruction-tuned for image reasoning and text generation from text and image inputs. We select LLaMA-3.2-Vision’s 11B version for its superior image reasoning ability, outperforming many multimodal models on several benchmarks.
3) \textbf{GPT-4o}~\cite{hurst2024gpt} is a large-scale VLM for general purposes, with an estimated 200B parameters.\footnote{The 200B parameter count is an unofficial estimate from~\cite{abacha2024medec}.} We include GPT-4o to compare our model with large-scale architecture.
It is worth noting that we exclude OutfitGAN~\cite{moosaei2022outfitgan} and DiFashion~\cite{xu2024diffusion} from our comparisons because they do not accept user text input, rendering them incompatible with our multimodal framework.

\begin{table*}[t]
  \centering
  \caption{Quantitative results of comparison with off-the-shelf VLM}
  \vspace{-3mm}
  \label{table:model-comparison}
  \resizebox{1.0\textwidth}{!}{
  \begin{tabular}{l|c|ccc|cccc|ccc}
    \hline
    \hline
    \multirow{2}{*}{Methods} & \multirow{2}{*}{Params} & \multicolumn{3}{c|}{Basic Recommendation} & \multicolumn{4}{c|}{Personalized Recommendation} & \multicolumn{3}{c}{Alternative Recommendation} \\
    & & S-BERT & CTS & CIS & S-BERT & CTS & CIS & Per. & S-BERT & CTS & CIS\\
    \hline
    Show-O & 2B & 51.82 & 24.71 & 61.98 & 44.96 & 24.85 & 62.08 & 65.68 & 60.60 & 25.98 & 65.94 \\
    LLaMA-3.2-vision & 11B & 63.22 & 24.59 & 76.36 & 71.83 & 27.19 & 79.05 & 83.60 & 59.46 & 24.14 & 74.80 \\
    GPT-4o & 200B & 66.03 & 25.10 & 74.85 & 76.61 & 28.06 & 78.86 & 83.09 & 67.06 & 25.66 & 75.38 \\
    FashionM3 & 2B & \textbf{72.69} & \textbf{26.51} & \textbf{80.37} & \textbf{78.54} & \textbf{28.88} & \textbf{80.51} & \textbf{84.08} & \textbf{74.99} & \textbf{27.54} & \textbf{79.98} \\
    \hline
    \hline
  \end{tabular}
  }
  \vspace{-4mm}
\end{table*}

\vspace{2mm}
\noindent\textbf{Evaluation Metrics.} 
1) \textbf{Params} represents the number of model parameters, measured in billions. 
2) \textbf{Sentence-BERT Similarity (S-BERT)} assesses the cosine similarity between the generated and ground-truth recommendation sentences. S-BERT~\cite{reimers2019sentence} is chosen over traditional LLM metrics like BLEU or ROUGE because it captures deep semantic relationships rather than relying on surface-level word overlap, which is critical for recommendation tasks where different wordings may convey similar meanings in context. 
3) \textbf{CLIP Text Similarity (CTS)} is computed as the cosine similarity between the CLIP embeddings of the generated text and the ground-truth image, assessing textual alignment with the target. 
4) \textbf{CLIP Image Score (CIS)} measures the cosine similarity between the CLIP features of the generated image and the ground-truth images. 
5) \textbf{Personalization (Per.)} measures how well recommendations align with user preferences. It first calculates the preference feature by averaging the CLIP image features of a user’s past interactions. Then, it computes the cosine similarity between this preference feature and embeddings of generated images.

\vspace{2mm}
\noindent\textbf{Implementation Details}
We fine-tuned FashionVLM based on the pretrained Show-O model~\cite{xie2024show}, which was not subjected to the LLaVA-tuning stage. FashionVLM accepts up to 381 input text tokens and generates images at a resolution of 512$\times$512 pixels. Instruction fine-tuning was performed on the training split of the proposed FashionRec dataset introduced in Section~\ref{sec:dataset}, with samples randomly drawn from three tasks. To enhance FashionVLM’s image-generation capabilities, we incorporated description pairs of fashion products during training. The model was trained on an A100 GPU for 46 hours, utilizing 400,000 recommendation prediction samples and 200,000 image generation samples. During inference, each method first generates recommendation text based on the input multimodal query. Subsequently, item descriptions are extracted from the recommendations to generate corresponding fashion product images. Except for the Show-O model, all methods leverage FashionVLM for image generation. In contrast, Show-O uses its own generation capabilities, allowing us to compare their image generation performance with FashionVLM.

\vspace{-2mm}
\subsection{Comparison with VLM Baselines}
We compare FashionVLM against several VLM baselines across multiple tasks and the quantitative results are presented in Table~\ref{table:model-comparison}, where all similarity metrics (S-BERT, CTS, CIS, and Per.) are scaled by multiplying by 100 for better readability. Below, we analyze the performance on those three tasks, highlighting key observations for each.
\begin{itemize}
    \item In the \textbf{Basic Recommendation} task, we evaluate performance by inputting full multi-round dialogues and analyzing the final response. As shown in Table~\ref{table:model-comparison}, FashionVLM achieves superior performance compared to baseline models. Our model records a S-BERT score of 72.69, surpassing Show-O by 20.87, LLaMA-3.2 by 9.47, and GPT-4o by 6.66 points. This substantial improvement over Show-O highlights the effectiveness of our fine-tuning strategy, which enhances semantic alignment between generated and ground-truth recommendations. Compared to LLaMA-3.2, FashionVLM’s higher S-BERT score reflects its stronger capability to capture deep semantic relationships, which is critical for recommendation. Despite having significantly fewer parameters than GPT-4o, FashionVLM outperforms it in both S-BERT and CIS score, demonstrating that our lightweight model delivers better recommendation quality.
    \item For the \textbf{Personalized Recommendation} task, FashionVLM continues to excel, achieving a S-BERT score of 78.54, outperforming Show-O by 33.58 points, LLaMA-3.2 by 6.71 points, and GPT-4o by 1.19 points. The remarkable improvement over Show-O highlights the impact of fine-tuning with the FashionRec dataset, enabling better alignment with user preferences. Against LLaMA-3.2, FashionVLM demonstrates consistent gains across metrics, with a CTS score of 28.88 (vs. 27.19, +1.69) and a CIS score of 80.51 (vs. 79.05, +1.46). The Personalization (Per.) metric further showcases FashionVLM’s strength, with a score of 84.08, closely trailing GPT-4o’s 85.51 but surpassing LLaMA-3.2’s 83.60 by 0.48 points. These results are promising, given FashionVLM’s compact 2B parameter size compared to GPT-4o’s 200B, indicating that our model achieves near-equivalent or superior personalization with significantly lower computational complexity.
    \item In the \textbf{Alternative Recommendation} task, FashionVLM achieves a S-BERT score of 74.99, which outperforms Show-O (60.60) by 14.39 points, LLaMA-3.2 (59.46) by 15.53 points, and GPT-4o (68.05) by 6.94 points. Compared to LLaMA-3.2, FashionVLM’s superior S-BERT score highlights its ability to generate semantically relevant alternatives, a key aspect of this task. Against GPT-4o, despite FashionVLM’s much smaller parameter count, it achieves a competitive CIS score of 79.98 compared to GPT-4o’s 77.28, alongside a higher S-BERT, underscoring its efficiency in delivering high-quality recommendations. Additionally, FashionVLM’s CTS score 27.54 surpasses all baselines, indicating better textual alignment with alternative suggestions.
\end{itemize}

Across three tasks, FashionVLM consistently outperforms Show-O, validating the effectiveness of our fine-tuning approach in enhancing both semantic recommendation quality and image generation alignment. Compared to LLaMA-3.2, FashionVLM exhibits notable improvements, particularly in S-BERT, which reflects its superior semantic understanding for recommendation tasks. Compared to GPT-4o, FashionVLM achieves competitive or better results despite its significantly smaller scale, highlighting its efficiency and robustness. The consistently lower CTS scores across all models can be attributed to the inclusion of reasoning, query summaries, or additional context in the generated recommendation texts, which reduces their direct similarity to target image embeddings. This effect is less pronounced in CIS scores, as image-based comparisons better capture the intended recommendation output.

\vspace{-2mm}
\subsection{Qualitative Evaluation}
We present a qualitative analysis to visually compare the proposed FashionVLM's recommendation performance with other baselines. All generated images in the results are produced using the same FashionVLM image generation capability, with the differences arising solely from the distinct recommendation content provided by each method.

\begin{figure}[t]
  \includegraphics[width=0.5\textwidth]{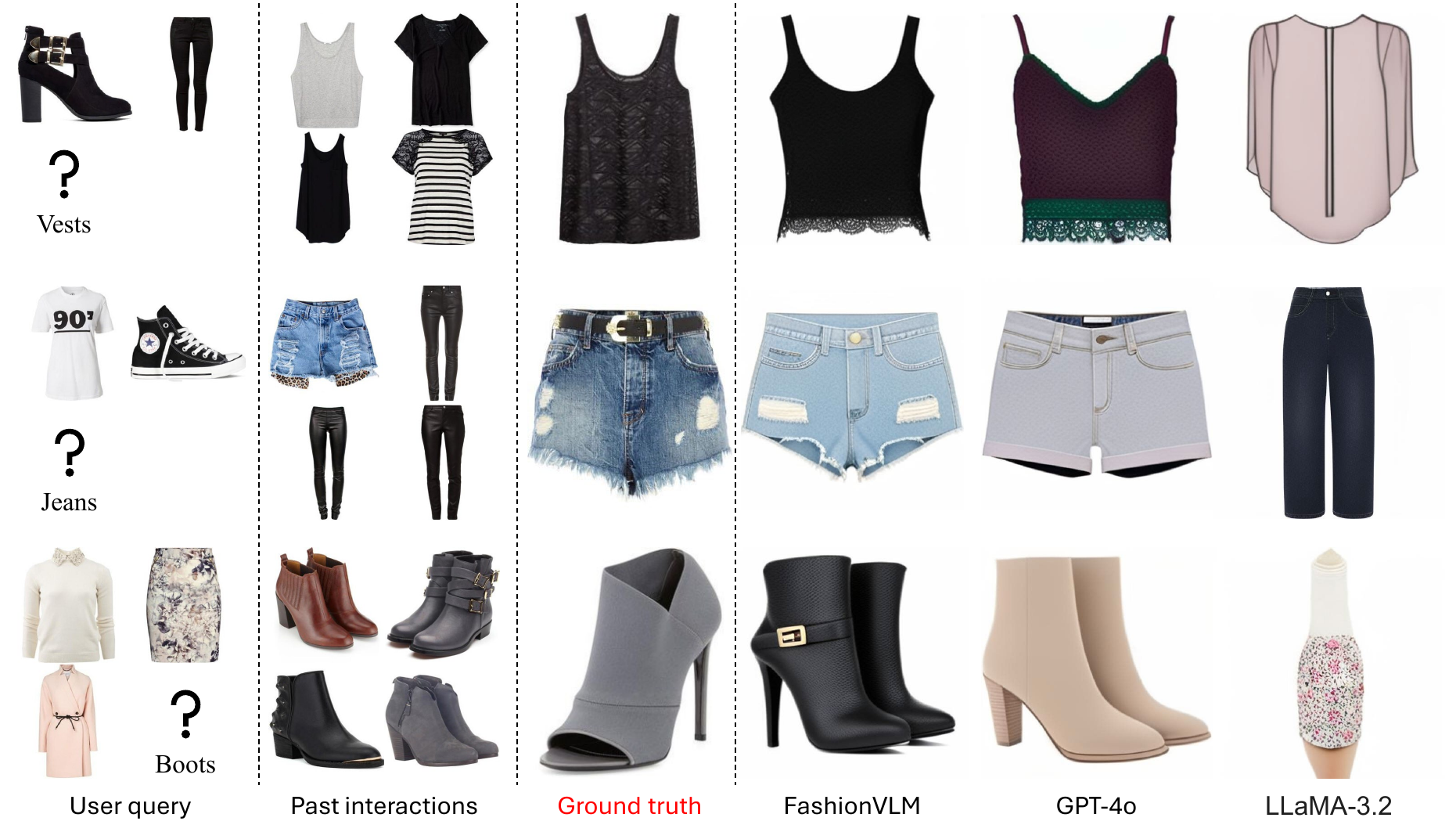}
  \vspace{-7mm}
  \caption{Qualitative results of generated images for personalized recommendation task.}
  \vspace{-5mm}
  \label{fig:case_study1}
\end{figure}

\noindent\textbf{Comparison on Personalization Task.} Fig.~\ref{fig:case_study1} showcases the generated product images from each method, given the user query and interaction histories. For the first example, the user wants recommendations for a vest. FashionVLM generates a black lace-trimmed crop top that mirrors the user’s historical interactions with dark, fitted tops, closely matching the ground-truth black textured vest. GPT-4o and LLaMA-3.2, however, produce a lace top and a pink blouse, respectively, which are less consistent with the user’s preference for darker, more structured vests. In the ``Jeans" category, FashionVLM recommends light-washed denim shorts, not only aligning with the user’s history of casual, distressed denim but also complementing the query’s outfit—a white T-shirt paired with black Converse sneakers—creating a cohesive and casual look, similar to the user selection. GPT-4o and LLaMA-3.2 suggest high-waisted jeans and dark wide-leg pants, which diverge from the user’s preference for shorter, more casual styles. These cases demonstrate FashionVLM’s superior capability to generate visually coherent recommendations aligned with users’ historical interactions.

\begin{figure}[t]
  \includegraphics[width=0.5\textwidth]{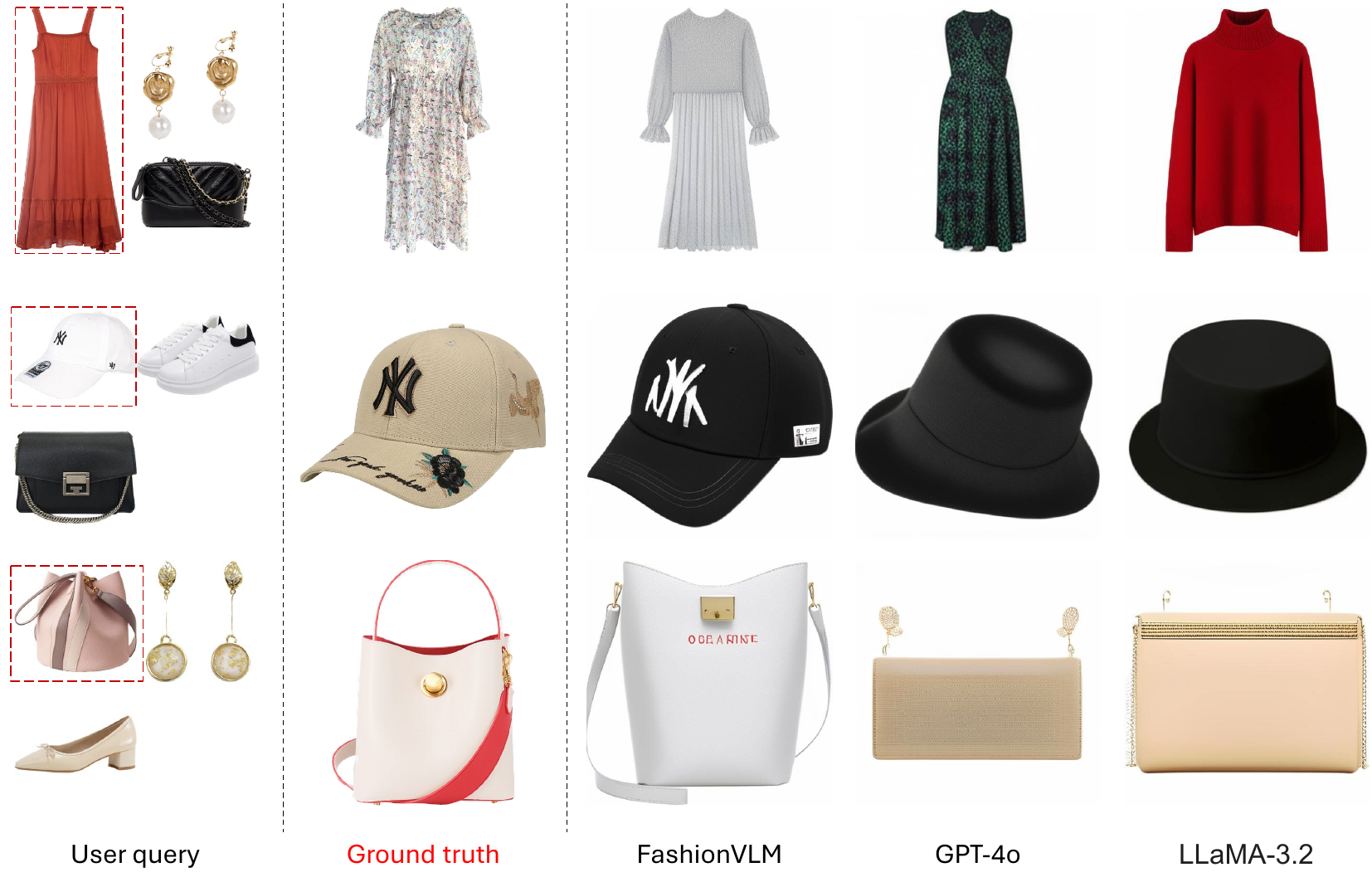}
  \vspace{-7mm}
  \caption{Qualitative results of generated images for alternative recommendation task.}
  \vspace{-6mm}
  \label{fig:case_study2}
\end{figure}

\noindent\textbf{Comparison on Alternative Task.} Fig.~\ref{fig:case_study2} presents the three examples for the Alternative Recommendation task. In each example, the user query section highlights the item the user wishes to replace, marked with a red box, while the Ground Truth represents the target alternative item. The three columns on the right display the alternative items generated by three methods.
For the first example, the user seeks an alternative to a red dress. The ground-truth item suggests a long-sleeved, floral-patterned dress. FashionVLM generates a long-sleeved dress in light gray, creating a harmonious and elegant ensemble. GPT-4o recommends a green-patterned dress, and LLaMA-3.2 produces a red turtleneck sweater, which captures some pattern or color but lacks the cohesive elegance of the ground truth.
The second and third examples demonstrate our advantages in accessory recommendations by accurately capturing the key features of the target products. For the second example, the user seeks an alternative to a white cap, with the ground truth suggesting a beige baseball cap. FashionVLM recommends a black baseball cap with a white `NY' logo on the front, aligning with the casual style. At the same time, GPT-4o and LLaMA-3.2 suggest a black fedora hat, deviating from the casual nature of the outfit. In the third example, the user aims to replace a bucket bag, with the ground truth providing a bucket bag, featuring a white and red color scheme. FashionVLM generates a white bucket bag featuring a gold clasp and a long strap, maintaining functionality and style, whereas GPT-4o and LLaMA-3.2 each recommend a beige leather pouch bag, which is misaligned with the ground truth. FashionVLM’s precise recommendations showcase its better understanding of item replacement.

\vspace{-3mm}
\subsection{User Study}
\subsubsection{Participants}
We conducted a one-hour remote user study with 12 participants (six females and six males, average age 30 years, standard deviation (SD)=5.06) to evaluate the effectiveness of FashionM3. In addition to the differentiation among their gender and age, the fashion knowledge of these participants varied (mean = 2.75, SD = 1.48 on a 5-point scale), as did their online shopping frequency (mean = 3.08, SD = 1.38 on a 5-point scale), representing a diverse range in fashion understanding and e-commerce engagement.

\subsubsection{Procedure}
\begin{itemize}
    \item \textbf{Introduction} (10 minutes). We began with a comprehensive demonstration of the FashionM3 to familiarize participants with its capabilities, utilizing the example described in Section~\ref{section_usage_scenario}. Subsequently, participants signed informed consent forms.
    \item \textbf{Occasion-Based Outfit Creation Task} (30 minutes). To evaluate FashionM3’s capabilities, we designed a task where participants aimed to curate an outfit for a self-defined occasion that matched their needs. Participants described their chosen scenario and used FashionM3 to generate, personalize, and visualize a suitable outfit, engaging with features like recommendation generation, style customization, alternative suggestions, image generation, and virtual try-on. To address the cold start, the moderator offered example scenarios if needed, but participants were encouraged to define their own. The task was completed within 30 minutes.
    \item \textbf{Questionnaire and Interview} (20 minutes). Upon completion of the styling phase, each participant completed a questionnaire of ten questions, evaluating the FashionM3's functionality and user experience. This questionnaire utilized a 5-point Likert scale and incorporated positively and negatively framed questions to minimize potential response bias. Subsequently, we conducted a semi-structured interview, during which participants discussed their experiences and offered suggestions for potential improvements.
\end{itemize}

\begin{figure*}
  \includegraphics[width=\textwidth]{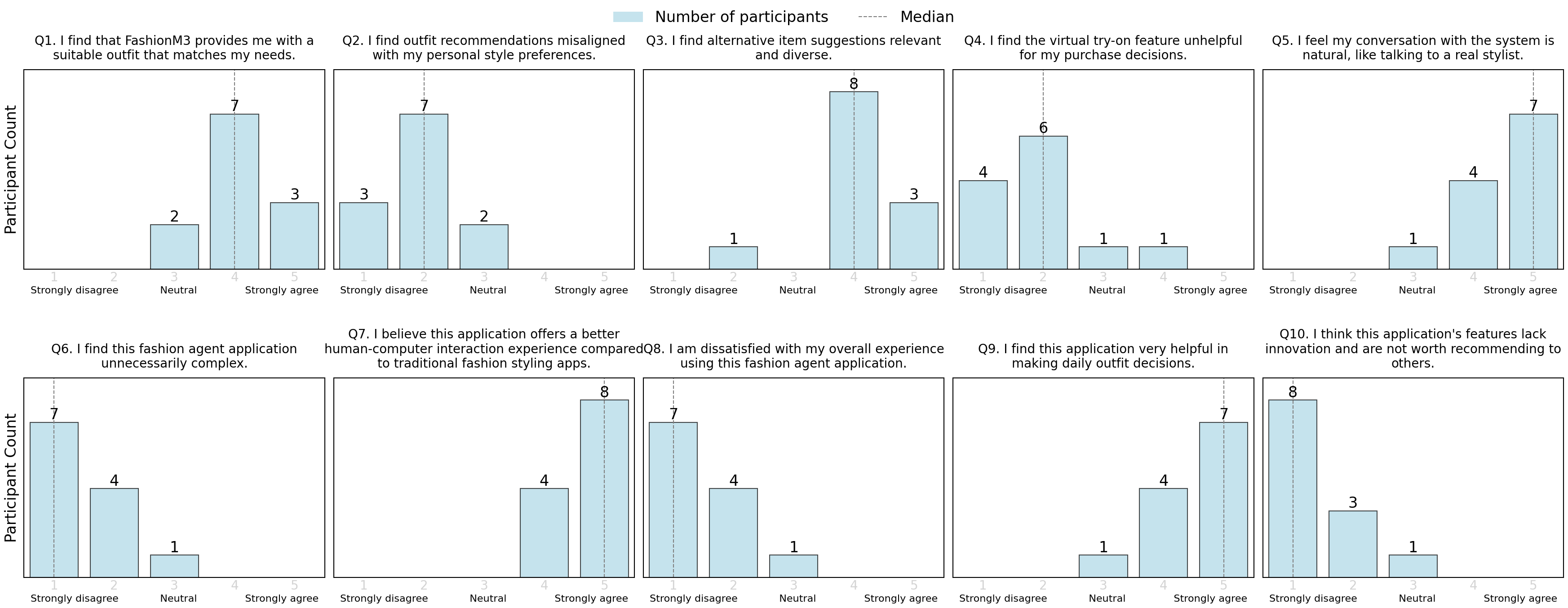}
  \vspace{-7mm}
  \caption{Statistical results of user experience survey (n=12). The survey comprises ten questions on a 5-point Likert scale, ranging from ``Strongly disagree" to ``Strongly agree". Bar heights represent the number of participants selecting each option.}
  \vspace{-5mm}
  \label{fig:user_rating}
\end{figure*}

\subsubsection{Results and Discussion}
We present participants' responses to FashionM3 and observations on their styling practices and interactions with the agent.
\begin{itemize}
    \item \textbf{Recommendation Effectiveness.} We evaluated FashionM3’s ability to assist users in creating an outfit for their chosen occasion, focusing on overall satisfaction, style alignment, and the quality of alternative suggestions. For overall satisfaction, Q1 responses showed that 83.3\% of participants agreed or strongly agreed that they were satisfied overall with FashionM3’s ability to help them create an outfit for their chosen occasion, suggesting the task was generally completed effectively. For style alignment, 75.0\% of participants disagreed or strongly disagreed with Q2, indicating that FashionM3’s recommendations were mostly consistent with their personal style preferences. P3 noted, ``It was about right for my style after I added some details." For alternative suggestions, 91.7\% of participants agreed or strongly agreed with Q3 that FashionM3 provided relevant and diverse alternative items, with P11 stating, ``The other options were decent, gave me something to work with." These findings suggest FashionM3 supports users in achieving their styling goals with recommendations that generally meet their needs.
    \item \textbf{User Experience.} A notable feature is the virtual try-on functionality, which aims to enhance users' ability to visualize recommended outfits. Survey results for Q4 (I find the virtual try-on feature \textbf{unhelpful} for my purchase decisions) showed a median response of ``Disagree" on a 5-point scale, suggesting that users generally found this feature helpful. One participant (P5) commented, ``The try-on feature lets me directly perceive how the outfit would look together." The multi-modal interaction and overall human-computer experience were also assessed. For both Q5 and Q7, the median responses were 5 out of 5. A participant (P8) observed: ``The conversation flows smoothly, allowing me to communicate through text, images, and dialogue easily." These elements appear to create an environment for exploring fashion possibilities, potentially offering a more experiential and personalized approach to online shopping.
    \item \textbf{Satisfaction and Style Inspiration.} The study evaluated the agent's overall user satisfaction. Results for Q9 showed a strong positive trend, with 91.67\% of participants finding this application helpful in making daily outfit decisions. As for Q10, all but one participant disagreed or strongly disagreed that the application's features \textbf{lack} innovation, suggesting a high recommendation potential. Qualitative feedback further supports these quantitative results, highlighting specific features contributing to user satisfaction and the system's inspirational value. One participant, P3, noted, ``It provides excellent explanations, which is particularly helpful for those without extensive fashion knowledge." Additionally, P2 praised the diversity of outfit recommendations: ``The recommended outfit variations are quite extensive," which exposes users to a wide range of fashion possibilities. Notably, the agent acts as a source of inspiration; as P4 remarked, ``Even if I don't fully adopt the outfit suggestions, they are still very inspirational." This approach encourages creativity and personal style development, enabling users to learn from and adapt recommendations to their preferences.
\end{itemize}

\vspace{-3mm}
\subsection{Usage Scenario}
\label{section_usage_scenario}
\begin{figure*}[t]
  \includegraphics[width=\textwidth]{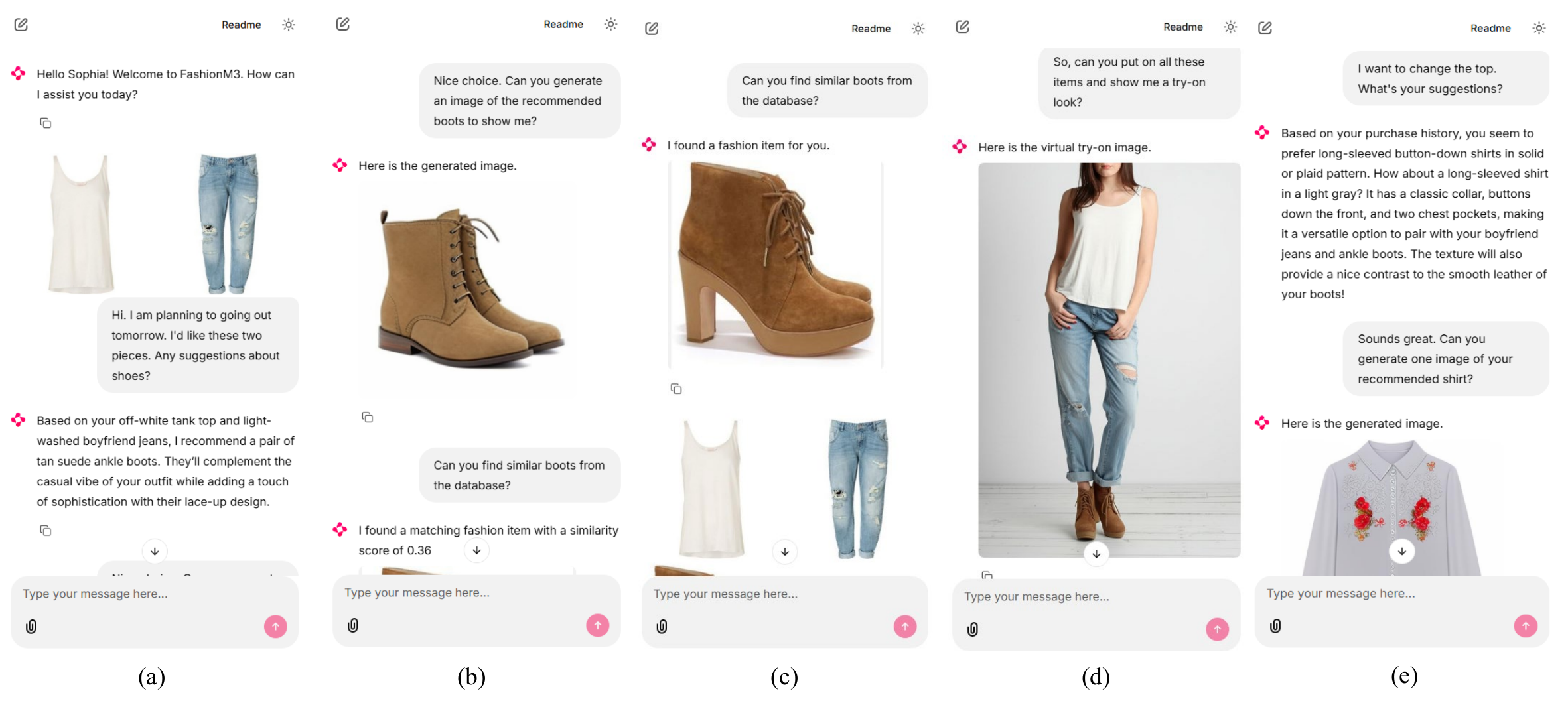}
  \vspace{-7mm}
  \caption{Interactive workflow of FashionM3 assisting Sophie in finding an outfit for a casual day out. The process unfolds in five steps (a-e). (a) Sophie’s initial request with uploaded images of a top and jeans, followed by the agent’s recommendation based on her historical data; (b) Generated image of the recommended tan suede ankle boots to facilitate Sophie’s visual perception of the recommendation; (c) Retrieval of similar boots from the database, showcasing the system’s product matching capability; (d) Virtual try-on image of a model wearing the complete recommended outfit, including the top, jeans, and boots; (e) Recommend alternative item of a long-sleeved shirt based on Sophie’s purchase history, with a generated image to help her visualize the updated look.}
  \vspace{-5mm}
  \label{fig:scenario1}
\end{figure*}

Suppose that Sophie (a user with ID ``115'' from the Polyvore-519 dataset, named for ease of understanding), who often struggles with fashion choices and outfit coordination, is preparing for a casual day out. After spending hours browsing online fashion retailers and trying various search combinations, Sophie couldn’t find an ensemble that truly matched her vision. Frustrated with her unproductive search, she decides to try out the proposed fashion agent, FashionM3, hoping for a more personalized and efficient recommendation experience. Sophie initiates the conversation through the chat interface (\textbf{Fig.~\ref{fig:scenario1}-a}). She types, \textit{``Hi. I am planning to go out tomorrow. I like these two pieces. Any suggestions about shoes?"} and uploads images of a top and a pair of jeans. After retrieving Sophie’s historical data from the database, the agent recognizes her preference for casual, minimalist styles and responds with a recommendation: \textit{``I recommend a pair of tan suede ankle boots. They’ll complement the casual vibe of your outfit while adding a touch of sophistication with their lace-up design."}

The agent then leverages its generative capabilities to enhance Sophie’s experience (\textbf{Fig.\ref{fig:scenario1}-b}). Based on the recommendation, the system generates an image of the suggested tan suede ankle boots, allowing Sophie to visualize the item directly. This visual representation strengthens her intuitive understanding of how the boots would pair with her chosen outfit. Curious about purchasable products, Sophie asks \textit{``Can you find similar boots from the database?"} The agent immediately taps into its product retrieval functionality (\textbf{Fig.\ref{fig:scenario1}-c}). It searches the fashion database and identifies a pair of boots, presenting them to Sophie with a detailed description: \textit{``I found a matching fashion item for you."} This retrieval of similar items demonstrate the system’s ability to match user requests with relevant products. If integrated with a sales system built on an MCP server, Sophie could directly order the boots, streamlining the purchasing process.

To further assist Sophie in visualizing the complete look, the agent employs its virtual try-on functionality (\textbf{Fig.\ref{fig:scenario1}-d}). It generates an image of a model wearing the recommended outfit—the white tank top, light-washed jeans, and tan suede ankle boots. This user-friendly feature allows Sophie to see how the pieces come together as a cohesive ensemble, helping her feel more confident in her choice. Finally, the agent showcases its alternative recommendation capabilities (\textbf{Fig.\ref{fig:scenario1}-e}). Sophie asks, \textit{``I want to change the top. What’s your suggestion?"} The system analyzes Sophie’s historical data, noting her preference for long-sleeved button-down shirts in solid or patterned designs. It recommends a light gray long-sleeved shirt with a classic collar and subtle floral embroidery, explaining: \textit{``Based on your purchase history, you seem to prefer long-sleeved button-down shirts in solid or light patterns. How about a long-sleeved shirt in a light gray? It has a classic collar, buttons down the front, and two chest pockets, making it a versatile pairing option with your boyfriend jeans and ankle boots. The texture will also provide a nice contrast to the smooth leather of your boots!"} The agent then generates an image of the recommended shirt, allowing Sophie to visualize the alternative top and appreciate its fit with the rest of her outfit. After examining the virtual try-on and alternative recommendation, Sophie feels the outfit aligns perfectly with her style and the casual nature of her day out. The personalized suggestions, combined with the visual aids, make her shopping experience efficient and enjoyable.

\vspace{-2mm}
\section{Limitations and Future Work}
While the agent utilizes VLM, its effectiveness is contingent on the quality and diversity of the underlying fashion database. The personalized recommendations may not accurately reflect user's prefence if the database lacks user interactions. Furthermore, the current agent operates inventory-agnostic when generating outfit recommendations. It may lead to a mismatch between recommendations and available items, particularly in scenarios with limited stock. The assistant's performance may vary based on the specificity of user input; vague requests could result in less tailored recommendations. Furthermore, while beneficial, the virtual try-on technology may not accurately account for individual body types and sizes. This limitation could lead to discrepancies between the visualized outfits and how they appear to users. Lastly, the agent operates within a predefined framework, which may not adapt dynamically to evolving fashion trends.

In future work, improvements can be made by implementing an inventory-aware recommendation algorithm, which can be achieved by injecting inventory data into the query. Enhancing the VTON model to incorporate advanced body scanning technologies could provide more accurate visualizations.

\vspace{-2mm}
\section{Conclusion}
This paper proposes FashionM3, a multimodal, multitask, and multiround fashion assistant built on the fine-tuned FashionVLM, enabling interactive styling experience. FashionM3 supports intuitive interactions through natural language, image inputs, and continuous dialogue, integrating visual analysis, user profiles, and a comprehensive fashion database to deliver tailored outfit suggestions that reflect individual preferences and contextual needs. The introduction of the FashionRec dataset, enables effective training of FashionVLM across diverse recommendation tasks. Both quantitative and qualitative evaluations demonstrate FashionVLM’s superior recommendation performance over baselines, while user studies highlight FashionM3’s practical value in providing engaging and personalized fashion advice. By leveraging multimodal inputs and iterative feedback, FashionM3 offers a more dynamic solution compared to existing methods.


\bibliographystyle{IEEEtran}
\bibliography{manuscript}

\newpage


\appendix
\subsection{System Prompts for Generating Dialogues}
\label{appendix:recommendation-based-queries}
\noindent\textbf\small{1. For Basic Recommendation.}
\begin{lstlisting}[language=]
messages=[{`role': `system', `content': ```
As a fashion expert, generate a user-system conversation for training a fashion stylist model. Your goal is to create natural, concise, and relevant dialogues based on the provided partial outfit and target items.

**Guidelines:**
- Create a "Basic Recommendation" conversation with a dynamic number of rounds. The number of rounds should be less than or equal to the number of target items, reflecting a progressive recommendation process. For example, if there are 3 target items, the conversation may have 1-3 rounds, as the user might ask for one item at a time or multiple items in a single question.
- The user's first question should include mimic how a user might naturally refer to an uploaded image.
- Ensure the user's questions collectively mention all desired categories for recommendations, covering every target item in the set by the end of the conversation. Categories can be general (e.g., "shoes" instead of "sneakers").
- Do not mention specific target items (e.g., "red sneakers") in the user's query. Use the target item details in the system response to justify the recommendation.
- Ensure questions connect the Partial Outfit and Target Items, avoiding generic queries.
- Vary system responses in tone, structure, and style for natural, helpful interactions.

IMPORTANT: The number of rounds should be less than or equal to the number of target items.
'''}]
\end{lstlisting}

\noindent\textbf\small{2. For Personalized Recommendation.}
\begin{lstlisting}[language=]
messages=[{`role': `system', `content': ```
Create a user-system conversation for training a personalized fashion stylist model. Focus on developing natural, concise, and relevant dialogues using the provided partial outfit, target items, and user's historical interacted items.

**Guidelines:**
- **Personalized Recommendation**: Develop dialogues that reflect user preferences through provided historical interacted items. Summarize these preferences and discreetly add them to the end of user queries, post-question, simulating backend database information injection.
- **User Queries**: The user's question should mimic a natural reference to an uploaded image. Use general item categories (e.g., "shoes" instead of specific items like "red sneakers"). The connection to Partial Outfit and Target Items should be clear, avoiding overly generic queries.
- **Post-Query Injection**: Add summarized user preferences based on historical interactions at the end of user's questions, maintaining the natural flow of conversation.
- **System Responses**: System responses should vary in tone, structure, and style, providing engaging and contextually relevant recommendations for each user inquiry. If historical preferences don't fully align with the Target Item, reflect this subtly in the response (e.g., uncertainty or an alternative suggestion).
- **Valid Flag**: Not all input samples are valid. Before setting the flag, evaluate if the user's historical interacted items align with the Target Item based on fashion attributes such as color, fit, design, or other style tags (excluding category). Set `valid` to 0 if none of the historical items share any relevant attributes with the Target Item; set it to 1 if at least one historical item matches in color, fit, design, or other fashion tags. Default to 0 when alignment is unclear or insufficient.

Ensure that there is only ONE round of conversation.
'''}]
\end{lstlisting}

\newpage

\noindent\textbf\small{3. For Alternative Recommendation.}
\begin{lstlisting}[language=]
messages=[{`role': `system', `content': ```
As a fashion expert, generate a user-system conversation for training a fashion stylist model. Your goal is to create a natural, concise, and relevant dialogue based on a given outfit and a specified changeable item that can replace one item in the given outfit.

**Guidelines:**
- Create an "Alternative Recommendation" conversation with exactly one round. The input consists of a complete outfit (Outfit A) with multiple items, each with a description, and a single changeable item (Item B) with its description, which is known to be a suitable replacement for a specific item in Outfit A (Item A). Item A and Item B are assumed to be replaceable because they belong to the same subcategory or have been paired with the other items in two outfits in the past.
- The user's question should mimic how a user might naturally refer to their current outfit, and express a desire to replace Item A with something different in the same category.
- The system response should recommend Item B as the replacement for Item A, using Item B's description to justify why it pairs well with the remaining items in Outfit A (excluding Item A) and fits the user's request.
- Ensure the user's questions collectively mention all desired categories for recommendations, covering every target item in the set by the end of the conversation. Categories can be general (e.g., "shoes" instead of "sneakers").
- Do not mention the Item B in the user's query. Use the category of Item A in the query (e.g., "sneakers") and the description of Item B in the system response.
- Ensure the question connects the user's current outfit to the replacement request, avoiding generic queries.
- The system response should be natural, helpful, and explain how Item B complements the remaining items in Outfit A.
- Vary user query and system responses in tone, structure, and style for natural, helpful interactions.

IMPORTANT: The conversation must have exactly one round.
'''}]
\end{lstlisting}

\newpage

\subsection{Examples of FashionRec Dataset}
\label{appendix:dataset-examples}
\begin{figure*}[t]
  \includegraphics[width=\textwidth]{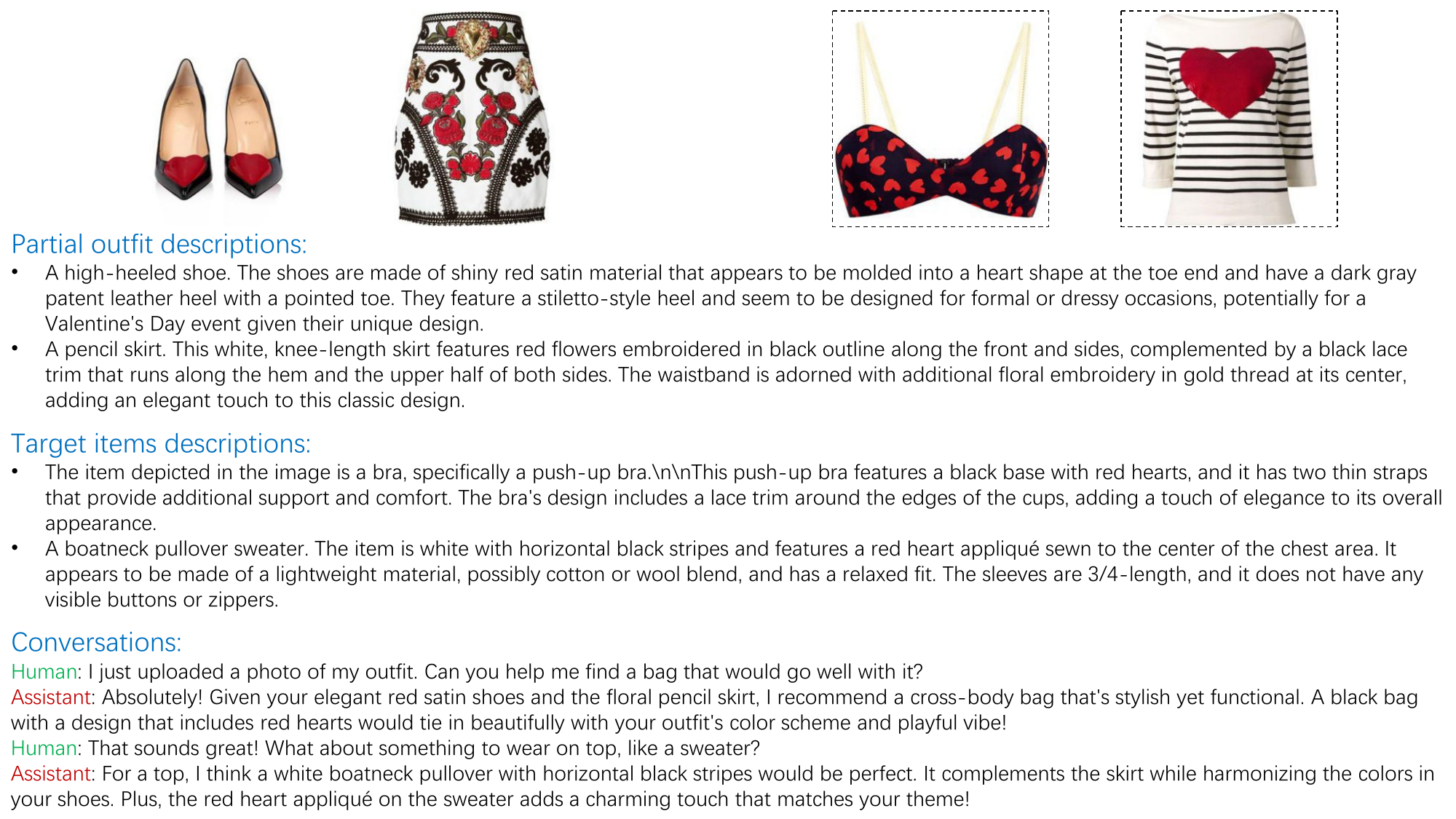}
  \caption{One example of basic recommendation task. The partial outfit is shown without a border, while the target items are indicated within a dashed border.}
\end{figure*}

\begin{figure*}[t]
  \includegraphics[width=\textwidth]{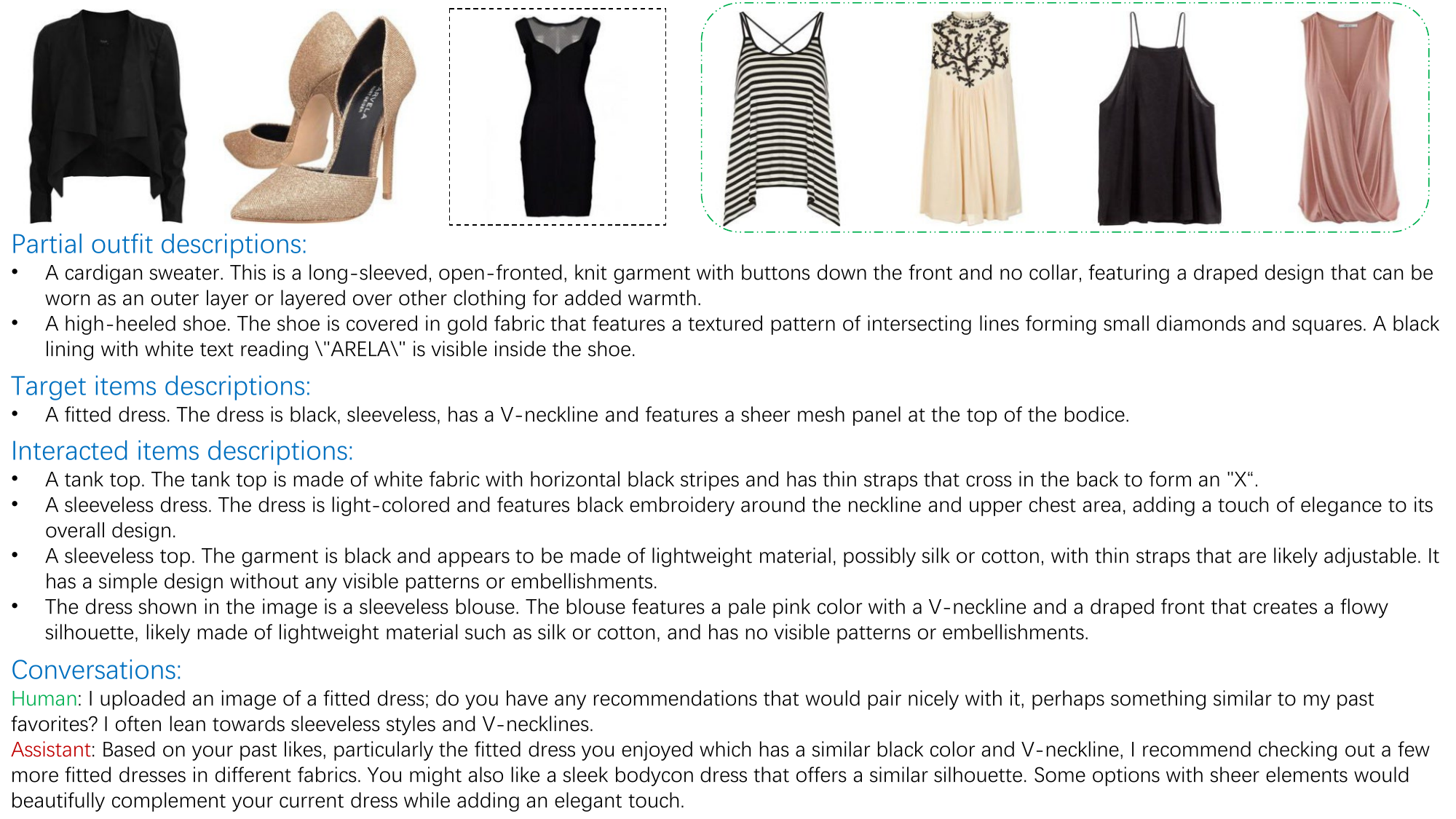}
  \caption{An example of the Personalized Recommendation task. The partial outfit is shown without a border, the target items are indicated within a dashed border, and the interacted items are enclosed in a green dashed border.}
\end{figure*}

\begin{figure*}[t]
  \includegraphics[width=\textwidth]{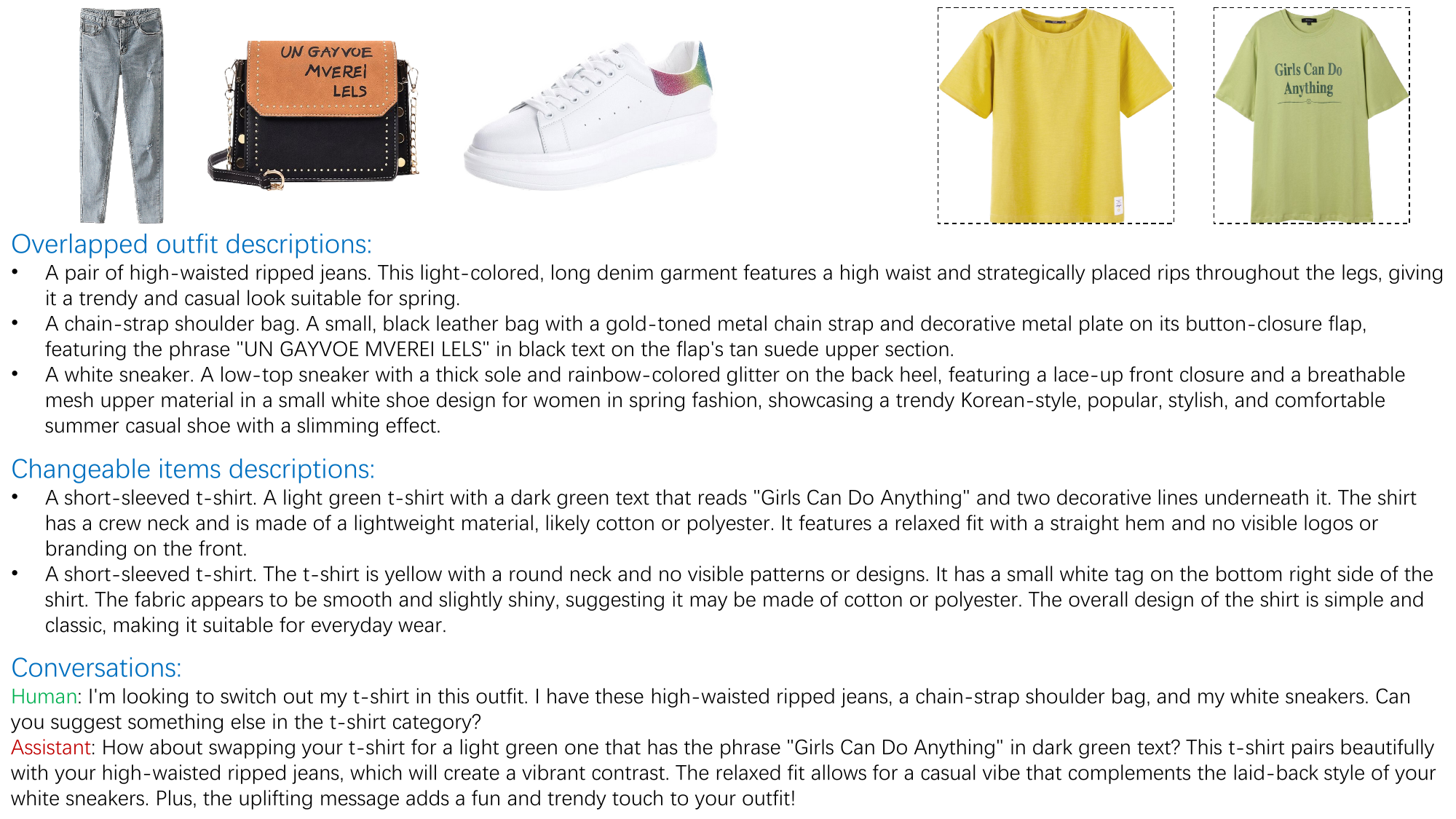}
  \caption{One example of the Alternative Recommendation task. The overlapped items are shown without a border, while the changeable items are indicated within a dashed border.}
\end{figure*}

\end{document}